\documentclass[twocolumn,wide]{adonis}
\usepackage[T1]{fontenc}
\usepackage{graphicx}
\usepackage[utf8]{inputenc}
\usepackage[linesnumbered, english, boxruled]{algorithm2e}
\usepackage{amsmath}
\usepackage[hidelinks]{hyperref}
\usepackage[nameinlink,capitalize]{cleveref}
\usepackage{xspace}
\usepackage{multirow}
\usepackage[caption=false]{subfig}
\usepackage[font=small]{caption}
\usepackage{siunitx}
\providecommand{\qty}[2]{\SI{#1}{#2}}
\usepackage{todonotes}
\usepackage[backend=biber]{biblatex}
\SetKwInput{KwInput}{Input}                
\SetKwInput{KwOutput}{Output}              
\newcommand\degree{\ensuremath{^\circ}\xspace}

\newcommand\eg{\mbox{e.\,g.}\xspace}

\title{Stand Up, NAO!}
\subtitle{Increasing the Reliability of Stand-Up Motions Through Error Compensation in Position Control}

\author{Philip Reichenberg and Tim Laue}

\affiliation{Universität Bremen, Fachbereich 3 -- Mathematik und Informatik, Postfach 330 440, 28334 Bremen, Germany}
\correspondence{s\_ksfo6n@uni-bremen.de,\\tlaue@uni-bremen.de}
\version{\today}

\abstract{
Stand-up motions are an indispensable part of humanoid robot soccer. A robot incapable of standing up by itself is removed from the game for some time.
In this paper, we present our stand-up motions for the NAO robot. Our approach dates back to 2019 and has been evaluated and slightly expanded over the past six years. We claim that the main reason for failed stand-up attempts are large errors in the executed joint positions.
By addressing such problems by either executing special motions to free up stuck limbs such as the arms, or by compensating large errors with other joints, we significantly increased the overall success rate of our stand-up routine. The motions presented in this paper are also used by several other teams in the Standard Platform League, which thereby achieve similar success rates, as shown in an analysis of videos from multiple tournaments.
}

\begin{document}

\maketitle

\section{Introduction}
\label{s:introduction}
In this paper, we present our approach for reliable stand-up motions on the NAO robot, functioning on a large spectrum of different aspects of wear and tear. The NAO robot is used by all teams in the RoboCup Standard Platform League (SPL). Here, the rules state that the robots need to successfully stand up by themselves, otherwise they become removed from the pitch for a given time. Therefore, having a working motion for such situations is crucial, especially since falling over -- for different reasons -- is very common as described in \cite{Winner-RoboCup-2025}.
The approach presented in this paper has been used by the team B-Human since 2019 \cite{Winner-RoboCup-2019} and did only undergo minor changes since then. During this period, it was also adopted by a number of other teams, which base their developments on different B-Human code releases \cite{GitHub-Naova-CodeRelease,GitHub-RZweiKickers-CodeRelease,GitHub-RedbackBots-CodeRelease,GitHub-RoboEireann-CodeRelease,GitHub-SPQR-Team-CodeRelease,GitHub-WisTex-United-CodeRelease}, which provides a significant basis for its evaluation.

The importance of research on this topic increased at RoboCup 2016. Here, for the first time, a subset of games was played on \qty{8}{\milli\metre} artificial turf instead of flat felt carpet. Since 2017, all SPL games are played on such a surface. This lead to a significant decline of the success rate of stand-up motions, as shown in \cref{t:getUps:past}. Even after two years, the analyzed teams still showed significant problems regarding their stand-up motions. Previously, the robots primarily failed their routines when something was broken or a mismatch in the kinematics occurred, like a stuck arm after a goal keeper jump. With the more challenging carpet, which offers a slightly less stable base for standing, the robots fell over more often while standing up. 

\begin{table*}[t]
	\caption[Success rate of stand-up motions of different leading teams in two competitions, before and after the switch to artificial turf, which lead to a significant percentage decrease of successful stand-up attempts.]{Success rate of stand-up motions of different leading teams\footnotemark[1] in two competitions, before and after the switch to artificial turf, which lead to a significant percentage decrease of successful stand-up attempts. More detailed data is provided at \cite{FallStatistics}.}\label{t:getUps:past}    
	\centering
	\begin{tabular}{r||c|c|c|c}
		& \multicolumn{2}{c|}{European Open 2016} & \multicolumn{2}{c}{RoboCup 2018}\\
		Team  & \#Tries & \%Success & \#Tries & \%Success \\
		\hline
		B-Human		& 48 & 89.6 & 166 & 76.5\\
		HTWK Robots	& 36 & 88.9 & 303 & 67.3\\
		Nao Devils 	& 106 & \textbf{94.3} & 258 & \textbf{78.3}\\
		rUNSWift	& & & 224 & 49.6\\
	\end{tabular}
\end{table*}

Furthermore, over the years, many SPL teams operate a growing number of older robots, which are subject to wear and tear, making them particularly susceptible to problems when getting up. This is an important aspect, which is addressed by the approach presented in this paper.
\footnotetext[1]{Due to bad or missing recordings, the statistics for HTWK Robots and Nao Devils only cover two games in 2016, while the B-Human statistics covers six played games.}

The remainder of this work is organized as follows: \cref{s:relatedWork} discusses related work. This is followed by an analysis of the causes for failed stand-up attempts in \cref{s:causes}. Our approaches for compensation, balancing as well as further improvements are described in \autoref{s:compensation} and \autoref{s:adjustments}. Subsequently, an overview of the architecture is given in \autoref{s:architecture}. Finally, \autoref{s:eval} presents an evaluation that covers several teams at multiple tournaments.

\section{Related Work}
\label{s:relatedWork}

A typical approach to creating robot motions is by using so-called keyframes. These are snapshots of robot poses, defining specific positions for the different joints for a single time frame. Smooth movements are created by the execution of an interpolated sequence of such keyframes. Currently, all SPL teams seem to base their stand-up motions on this approach. Apart from the teams that use B-Human code, as described in \autoref{s:introduction}, the code releases of a number of other teams include this kind of implementation \cite{GitHub-BerlinUnited-CodeRelease,GitHub-Hulks-CodeRelease,GitHub-NomadZ-CodeRelease,GitHub-rUNSWift-CodeRelease}. 

For the NAO robot, Kleingarn and Brämer \cite{Stand-Up-Devils-2024} propose a decision-tree approach, in which the stand-up routine is split in different segments and -- based on heuristics -- the robots use the segment which is assumed to result in the most successful stand-up try. Their results are included in \autoref{s:eval}. 
The team B-Human used a similar approach in 2018 \cite{GitHub-BHuman-CodeRelease2018}. Instead of using different segments, each robot could be given own alternative keyframes during the stand-up routine. The idea was to calibrate the stand-up motions for each robot when necessary, while also reducing the number of configuration files to a minimum. As this approach did not work out well, as documented in \autoref{t:getUps:past}, a solution that works on all NAO robots without additional calibration or configuration afterwards became more desirable.

Apart from the NAO robot, other humanoid robots need to stand up, too. A common approach is to apply Reinforcement Learning and to learn a policy in simulation, which is later transferred to a real robot. This principle has been successfully demonstrated by \cite{jeong2016learning} as well as by \cite{haarnoja2024learning}. For the latter, the stand-up was only one part of a learned overall football behavior. Such an approach is not yet known for the NAO robot. However, for many self-built or modifiable robots, which includes those used in the aforementioned works, the task of standing up is easier than for the NAO robot. This is because one has, to a certain extent, control over the kinematics, mass distribution, and motor power. Of course, when competing in a RoboCup Humanoid League competition, certain rules apply \cite{HumanoidRules2025}. Nevertheless, for instance, the maximum allowed length of the arms is larger than for the NAO, allowing for more simplistic stand-up approaches.

Furthermore, recent works by \cite{he2025learning} and \cite{wang2025booster} show promising results with Unitree G1 and Booster T1 robots, which are also used by many teams in the RoboCup now. In \cite{he2025learning}, two phases are used to learn a robust get-up routine. In the first stage, a policy is trained to learn rolling over or standing up, without deployment constraints, as this policy is not meant to be used on the real robot. In the second phase, another policy is learned to mimic the first one slowly and with more regularization. The whole simulation and learning process is done in IsaacGym, which allows for fast parallel simulations on Nvidia GPUs.
On the other hand, \cite{wang2025booster} is not directly about learning policies for standing up, but about their learning framework to learn fast and stable walks. They also use IsaacGym combined with domain randomization and reward shaping. 

Last but not least, \cite{BitBits-StandUp2022} presented an approach of using quintic splines, a PD-controller and a parameter optimization with a Sim2Real transfer. As explained in our analysis of failed attempts in \autoref{s:causes}, this approach is not suitable for now, due to the large occurring errors between the measured and requested joint positions on the NAO robot.

\begin{figure*}[t]
	\includegraphics[width=\textwidth]{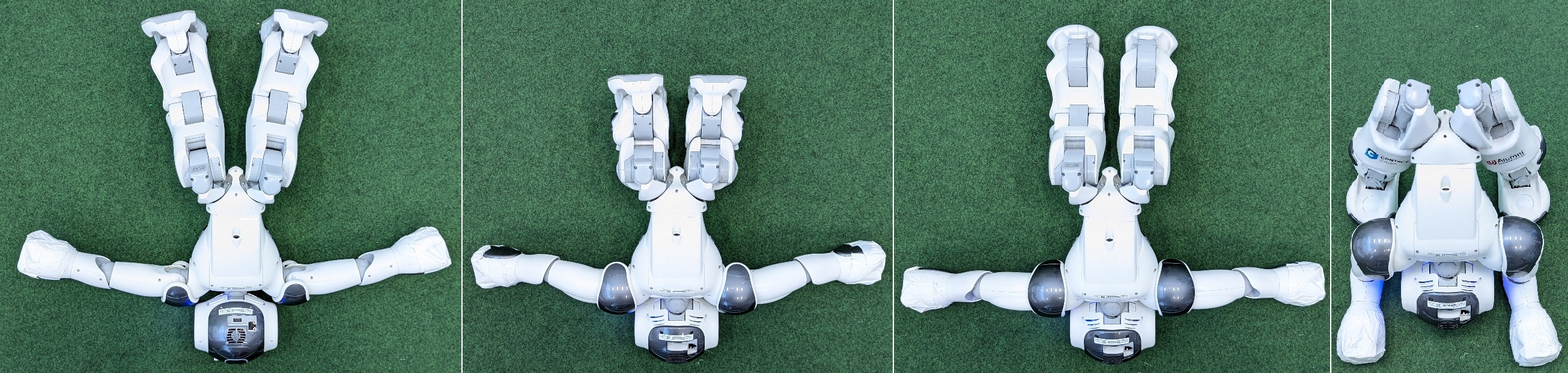}\\
	\includegraphics[width=\textwidth]{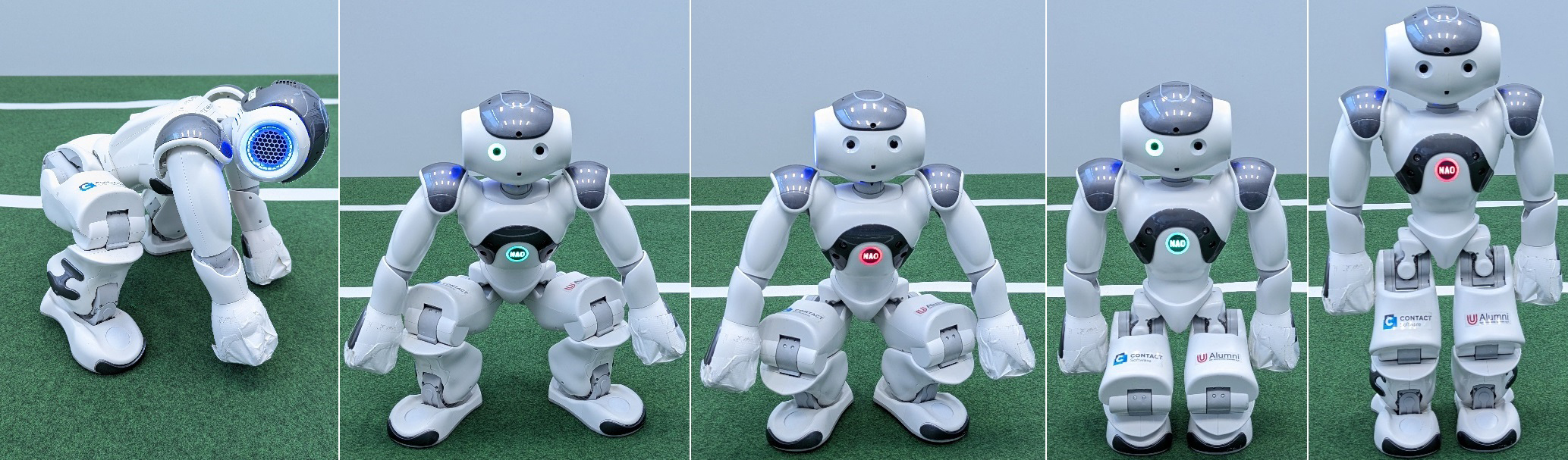}
	\caption{The front stand-up routine. Upper row: The robot starts with a default pose, pulls it legs together and rotates the arms, then first stretches the legs to lift itself up on the arms and head and then stretches the legs fully. This brings the robot in an ideal pose supported by the soles and arms with some momentum towards the legs, which is necessary afterwards to tilt the torso upwards. Bottom row: The robot shifts its CoM into the supporting area of its feet. Afterwards, the torso is tilted upwards, the legs aligned to shift the mass of the robot evenly and the legs are pulled together.}
	\label{f:getup:front:1}
\end{figure*}

\begin{figure*}[t]
	\includegraphics[width=\textwidth]{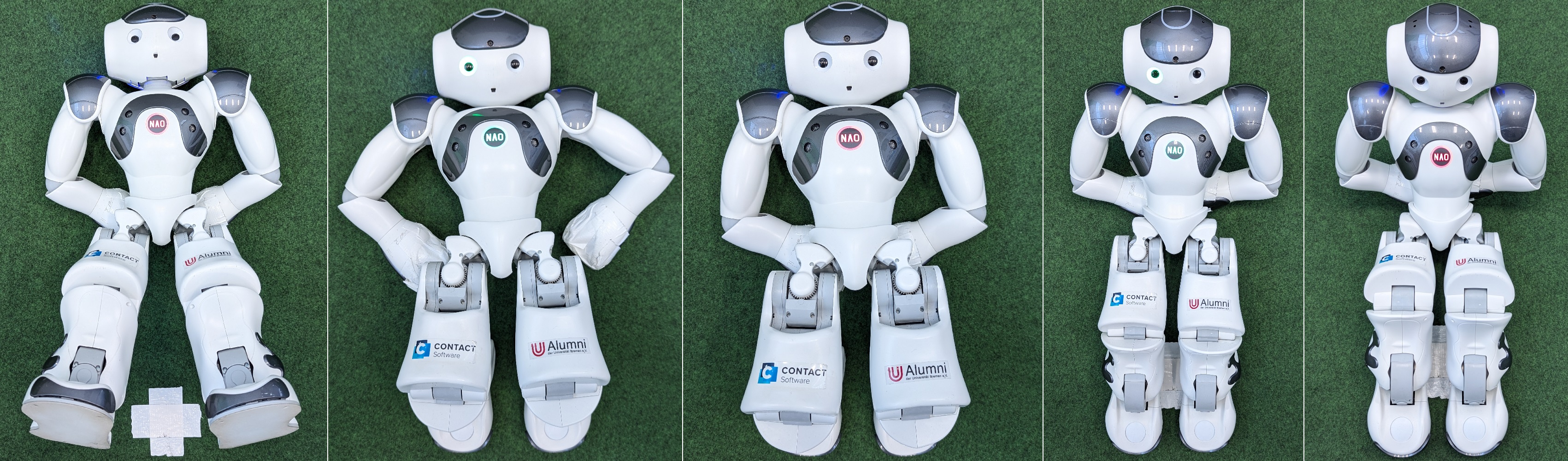} \\
	\includegraphics[width=\textwidth]{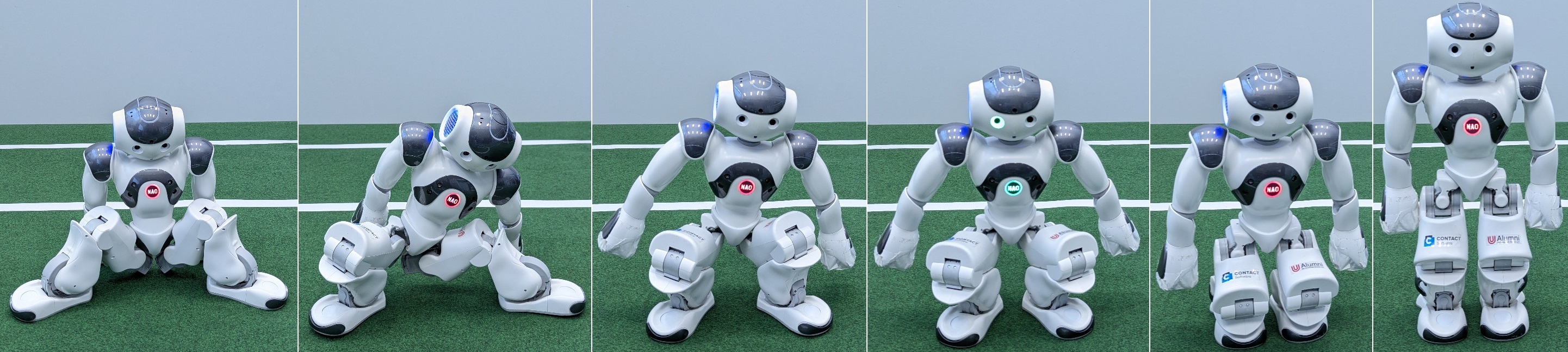}
	\caption{The back stand-up routine. Upper row: The robot moves the arms behind its back and then moves the legs upwards and forward to tilt the torso upright, by using the arms as a tilting point for the CoM. Bottom row: After tilting upright, the legs are spread apart and the arms are used to stabilize and prevent a fall backwards. Then the robot shifts itself on one leg followed by the other one. Afterwards, the legs are aligned to shift the robot mass evenly, which allows to pull the legs together and bring the robot safely into a standing pose.}
	\label{f:getup:back:1}
\end{figure*}

\section{Causes of Failed Attempts to Stand Up}
\label{s:causes}
Our approach considers two starting situations for carrying out a stand-up motion: lying on the front and lying on the back. For each of these two cases, a specific stand-up motion was developed. Both movements are shown in \autoref{f:getup:front:1} and \autoref{f:getup:back:1} respectively, slow motion videos are provided online\footnote{Slow motion video of the front stand-up: \url{https://youtu.be/IrvGDhXbcEw}, slow motion video of the back stand-up: \url{https://youtu.be/ZIa7yWbfOD4}}. In the rare case of lying on the side, a specific recovery motion is executed that turns the robot into one of the two starting situations.

During all games, every B-Human robot logs -- among many other things -- sensor data and requested joint positions of every execution cycle. This data provides a valuable source for analyzing failed attempts to stand up by comparing what a robot was trying to do to the actual states of its motors.
Note that for the NAO robot, only requests for desired joint positions can be set and not some torques or control parameters. Overall, we identified two main sources of failure:

\subsection{Stuck Arms}
During a competition game, other robots can walk or fall into another robot which is trying to stand up. Humans may intervene but it is very unlikely that such a disturbance is prevented before happening. Furthermore, due to bad or risky falls like goal keeper jumps, the arms might lay under the robot's own body. Such situations can influence the stand-up routine in a way that the arms are getting stuck. This is a severe problem, as NAO stand-up motions in general require the arms in specific positions. This is due the mass distribution of the robot, which requires a motion in which the robot has to tilt over at some point to bring its center of mass (CoM) into a supporting area, spanned by the soles and partially expanded by the arms.

\subsection{Stuck Joints}
Most fails result from one joint lagging behind or even being stuck, followed shortly after by other joints showing the same problem.
Typical balancing approaches, which can handle some variation of errors, are not sufficient anymore, if one joint shows a deviation of over \qty{20}{\degree}, which occurs in practice.

\begin{figure}
	\includegraphics[width=\columnwidth]{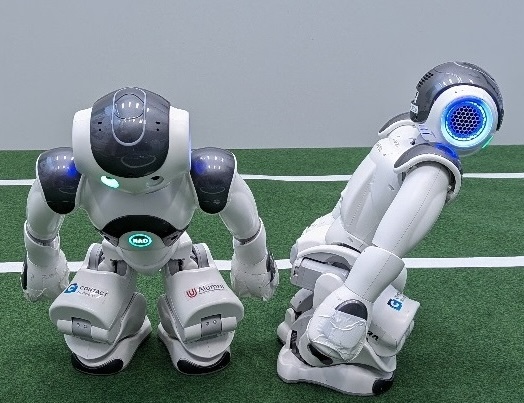}
	\caption{A previously stuck hip yaw pitch joint of \qty{30}{\degree} lets the robot tilt forward during standing up.}
	\label{f:compensation:hyp:motion}
\end{figure}

In most cases, one specific joint is the problem: the hip yaw pitch, which is a special feature of the NAO robot. It rotates both legs at the same time around the z- and y-axis plane. As a result, if this joint shows deviations, it will misplace both legs in their translation as well as in their rotation. During the stand-up, at some point, the legs are spread far apart, which converts to a large positive value for the hip yaw pitch. Afterwards, the legs need to move together again to reach some kind of sitting pose, which converts to a near zero value for the hip yaw pitch (see \autoref{f:getup:front:1} and \autoref{f:getup:back:1}). Due to this kinematic restriction of the NAO robot, if the hip yaw pitch changes, all other leg joints need to be adjusted as well to keep the same poses for the legs except for the z-axis rotation. But if the hip yaw pitch joint is stuck, the other leg joints will still move based on the planned motion and result in a movement, in which the robot is tilting forward and will fall over, as shown in \autoref{f:compensation:hyp:motion}.

\section{Compensation}
\label{s:compensation}

To handle the problem of joints getting stuck or reaching their target position slower than intended, we apply a compensation across multiple other joints. That way, if, for example, one leg joint should have rotated the robot backwards but is currently stuck due to insufficient strength, other joints compensate such error to bring the robot upright and prevent any unintended motions. To do so, we apply the error $\Delta_{j}$ of a given joint $j$ to other joints $\gamma$ (scaled by a given factor $p_\gamma$). Thereby, for each joint $\gamma$ in the set $J$, the original target position $R_{\gamma}$ becomes replaced by a modified target position $\hat{R}_{\gamma}$:

\begin{eqnarray}
	\forall \gamma \in \mathit{J}: \hat{R}_{\gamma} =& R_{\gamma} + \Delta_{j} \cdot p_{\gamma} \label{a:compensation:correction}
\end{eqnarray}

Those modified target positions $\hat{R}_{\gamma}$ are then used to interpolate between the start and end positions of the current keyframe execution. This means that for example after \qty{10}{\percent} execution time of the keyframe only \qty{10}{\percent} of the error correction is applied.

To determine the applied error $\Delta_{j}$, we first calculate a raw error $\delta_{j}$ based on the difference between the requested $r_{j}$ and the measured $m_{j}$ position of joint $j$:

\begin{eqnarray}
	\delta_{j} =& r_{j} - m_{j} \label{a:compensation:delta}
\end{eqnarray}

As the NAO robot has a motor delay of about \qty{36}{\milli\second} \cite{WalkStepAdjustment-RoboCup-2021}, we use the request $r_{j}$ from a past execution cycle, which we expect to be executed now. 

Once the stuck joint starts moving again, this compensation needs to be removed. For this, we calculate a simple prediction using the measured position change and multiplying it by 3, corresponding to the delay of the robot of about \qty{36}{\milli\second}, which corresponds to 3 execution cycles. This gives us a predicted error $\hat{\delta}_{j}$:

\begin{eqnarray}
	\hat{\delta}_{j} =& 3 \cdot (\delta_{{j,t}} - \delta_{{j,t-1}}) + \delta_{{j,t}} \label{a:compensation:prediction:delta}
\end{eqnarray}

with $\delta_{{j,t}}$ being the raw error of the current time frame $t$ and $\delta_{{j,t-t}}$ the raw error of the previous time frame.

To compute $\Delta_{j}$, $\hat{\delta}_{j}$ is further filtered by using the raw error $\delta_{j}$. Due to the simplistic prediction we do not want to determine a gain in the error, only a reduction. Therefore, the value closest to zero is used (see \autoref{a:compensation:error}). This is done by comparison. If the raw error $\delta_{j}$ and the predicted one $\hat{\delta}_{j}$ have different signs, we assume an error of zero. If the predicted one has an absolute lower value, it is used instead. Otherwise, the raw error value $\delta_{{j,t}}$ is used. The complete error calculation is as follows:

\begin{eqnarray}
	\Delta_{j} = & \normalbaselines\begin{cases}
		0 &, \hat{\delta}_{j} \cdot \delta_{{j,t}} \leq 0 \\
		\hat{\delta}_{j} &, |\hat{\delta}_{j}| < |\delta_{{j,t}}|  \\
		\delta_{{j,t}} &, \text{ otherwise} \label{a:compensation:error}\\
	\end{cases}
\end{eqnarray}

The compensation is always applied for the target position $R$. Once a new keyframe begins, this target position becomes the next starting position. To apply the compensation in a correct manner, it is applied for the current target position $R_0$ based on the current keyframe parameters and for the start position, which corresponds to the previous target position $R_{-1}$, based on the previous keyframe parameters. This way, once a stuck joint is moving again, no overcompensation is done. This approach also allows for different compensation parameters like which joints $\gamma \in \mathit{J}$ to use and by which factor $p_\gamma$ (see \autoref{a:compensation:correction}), while ensuring a smooth interpolation between both.

\begin{figure}[t]
	\subfloat[Without error compensation]{\includegraphics[width=\columnwidth]{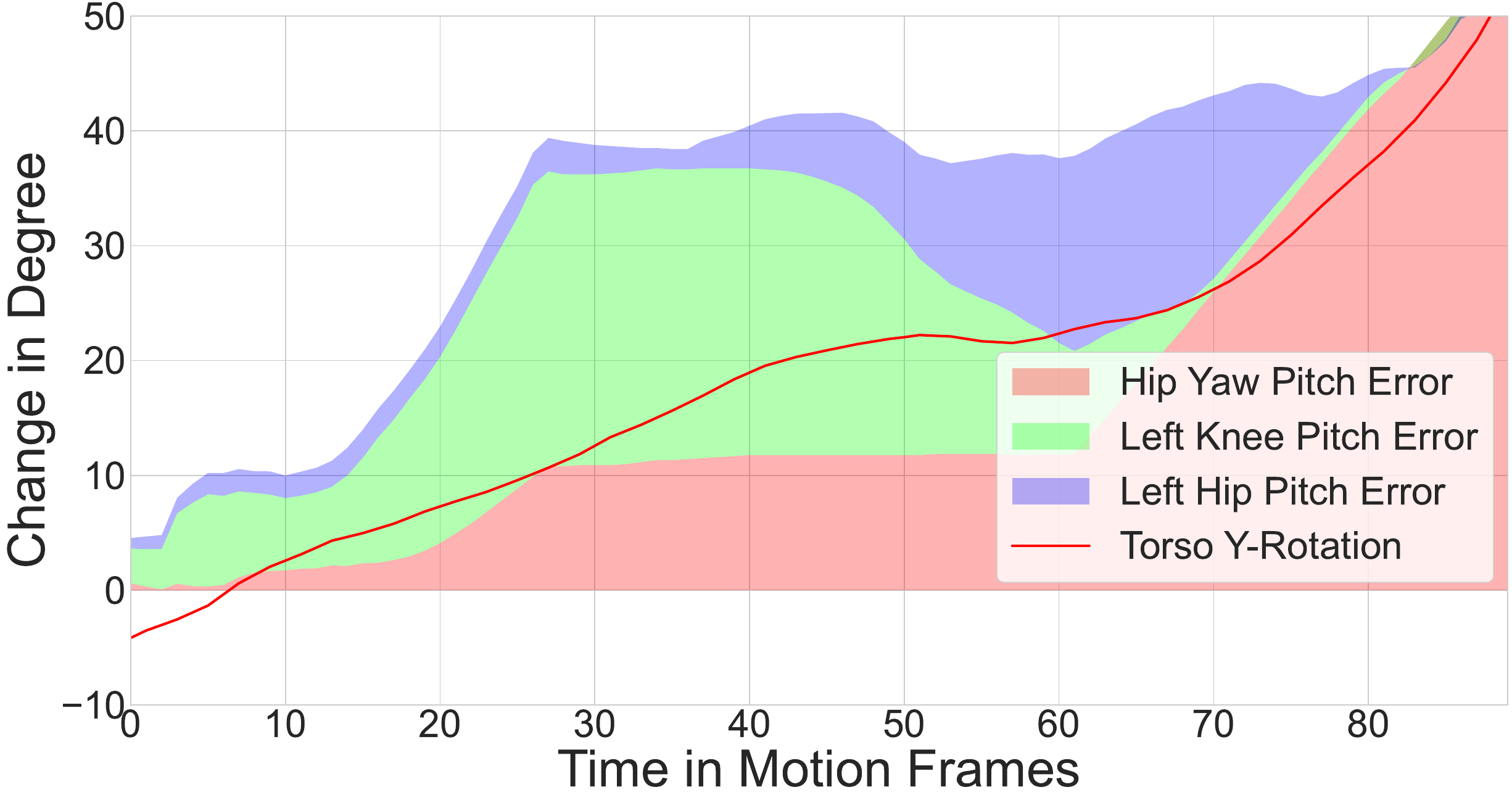}}\hfill
	\subfloat[With error compensation]{\includegraphics[width=\columnwidth]{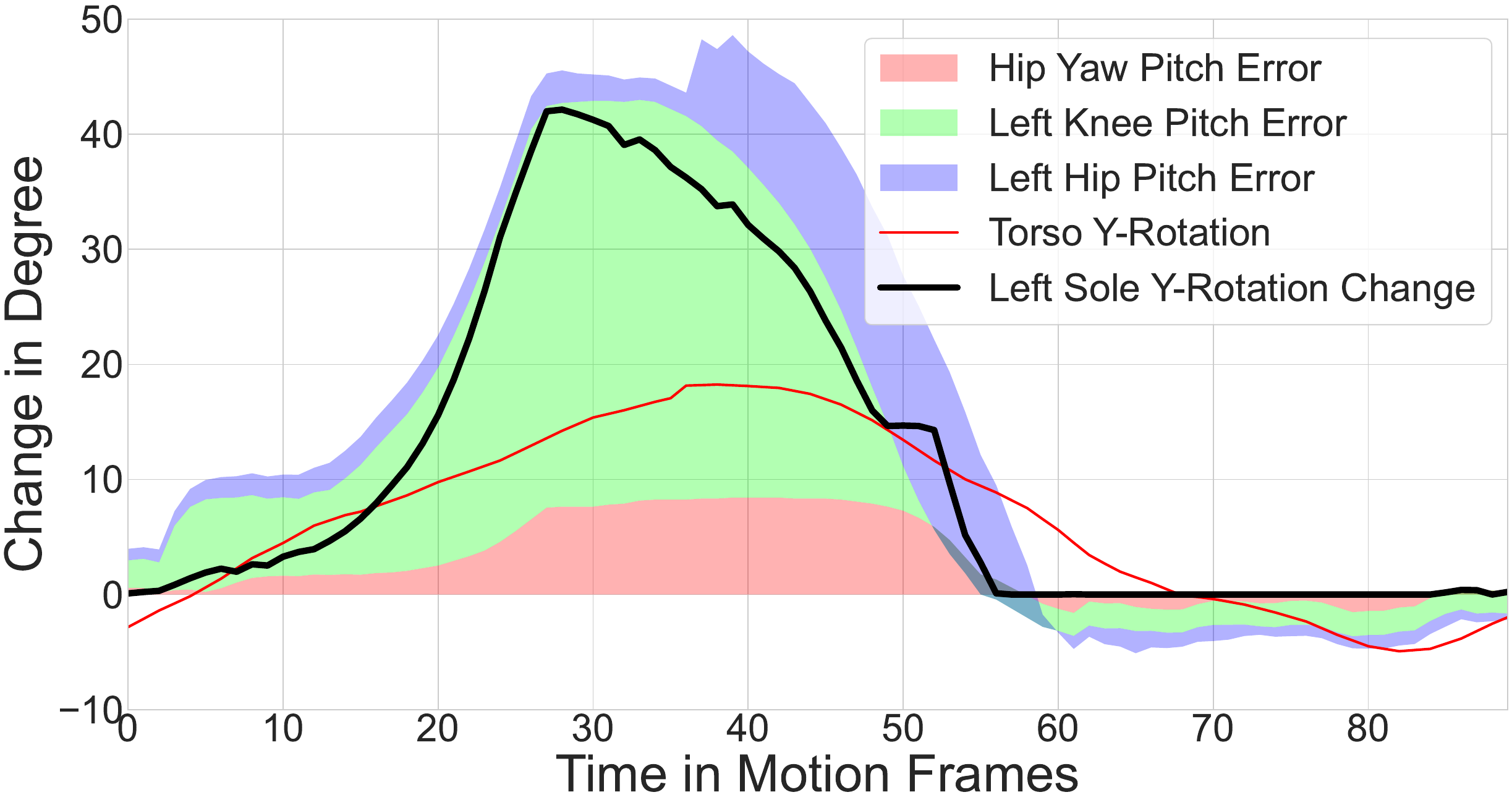}}
	\caption{A cutout of a stand-up attempt without and with the error compensation: a) The joints keep getting stuck or lagging behind, resulting in a fall. b) The left sole joints are used to compensate the errors, which keeps the torso stable and the joints free up after some time.}\label{f:compensation:plot}
\end{figure}

A comparison between a stand-up try without and with the proposed compensation is shown in \autoref{f:compensation:plot}. The robot's task was to stand up from lying on its back. After reaching an upright pose with the legs spread apart (see \autoref{f:getup:back:1}, second half), the robot needs to pull its legs together one after another. Previously, one joint would start lagging behind, with other joints following shortly after, resulting in the robot to fall over to the front. Now, even though the problem still occurs, other joints are used to compensate (see \autoref{a:compensation:correction}) for those errors. This is shown in the y-rotation change of the left sole, which displays that its rotation compared to the original planned one is significantly higher to compensate for the errors. The sole actively presses against the ground and holds the robot upright instead of ignoring the errors and falling forward. This is enough to keep the robot stable and allows the errors to fix themselves after some time, without the need to pause the stand-up routine.

Unlike to \cite{Stand-Up-Devils-2024}, the joints used for the compensation are not determined automatically. Every keyframe can define which joints to compensate and which other joints shall be used to do so and by which factor. An automatic process based on comparing the planned sole rotation and position and applying a correction based on those was tested and resulted in motions that were partially a lot more stable, but were also unable to release the stuck joints and reduce the errors and therefore resulted in significantly more failed stand-up tries.

\section{Further Extensions}
\label{s:adjustments}
The previously described compensation and balancing approaches are helpful to increase the stability of an already working motion, but cannot change a non-functional movement into a successful one. Also given enough execution time, a robot could theoretically stand up on any sensible surface starting from any pose. But as in RoboCup a fast stand-up routine is crucial, there is only a limited number of applicable options. This is why we introduced further extensions.

\subsection{Balancing}
As the robots need to execute the predefined motions on an slightly uneven underground with different state of wear of the joints, some errors will inevitably occur in the joint positions and in the CoM trajectory, in addition to joints getting stuck or moving slower than intended.

To handle both problems, we apply an exponential PD-balancer, using the CoM as a reference. The proportional part uses the difference between a configured expected CoM position and the measured one. The differential part uses the changes of the proportional one. The expected values are configured based on stable successful stand-up attempts that used no balancing. Here, we use the recorded CoM at the end of a keyframe as a reference and use a linear interpolation in between.

\subsection{Checks}
We added checks for the arms and the torso orientation in specific keyframes. If the robot is lying on its own arms, \eg after a goalkeeper jump, it makes no sense to continue the motion. Instead, those problems are detected by comparing the requested and measured joint positions of those limbs of interest after one specific keyframe and either a special motion is inserted to free up the limbs, or the stand-up routine returns to a previous keyframe to try again. In case such obvious problems are not detectable, we simply use the robot orientation and confirm whether the robot is still in a stable state and if not, the robot executes a fall motion with low stiffness and retries from the beginning. For this procedure, every keyframe has defined angle ranges for the torso orientation. This approach was also adapted by \cite{Stand-Up-Devils-2024}.
Those adjustments help reducing possible damages: no more joints or gears of our robots were broken during stand-up motions since 2018, which previously occurred regularly.

\subsection{Waiting}
In addition, we added a waiting feature. The idea is that in the transition phases from a lying pose to an upright pose supported by the soles, the robot executes a motion which tilts over the torso and moves the CoM into the supporting area of the soles. Due to different influences, like an uneven underground, joints not moving as expected or other robots walking into us, this tiling over motion could be delayed or slowed down. If the follow-up keyframe starts too early, the robot will fail to tilt over and fall back again. To handle this problem, we simply added a conditional wait time at the end of the responsible keyframes. To prevent a frozen motion, resulting from an unattainable torso orientation, the wait time is limited. This approach was also adapted by \cite{Stand-Up-Devils-2024}.

\subsection{Ankle Roll Oscillation}
Furthermore, since 2025, we added an explicit oscillation. Once the robot reaches a pose supported by the soles and the balancing is active, the ankle rolls are changed by a small value of up to \qty{1}{\degree}. This value is adjusted by a sine curve over a phase time of \qty{200}{\milli\second}. This helps the robot to further prevent soles stuck in the ground. The idea was tested on a \qty{38}{\milli\metre} artificial turf field, on which the robot could move the legs more easily together than without this explicit oscillation.

\begin{figure*}[t]
	\includegraphics[width=\textwidth]{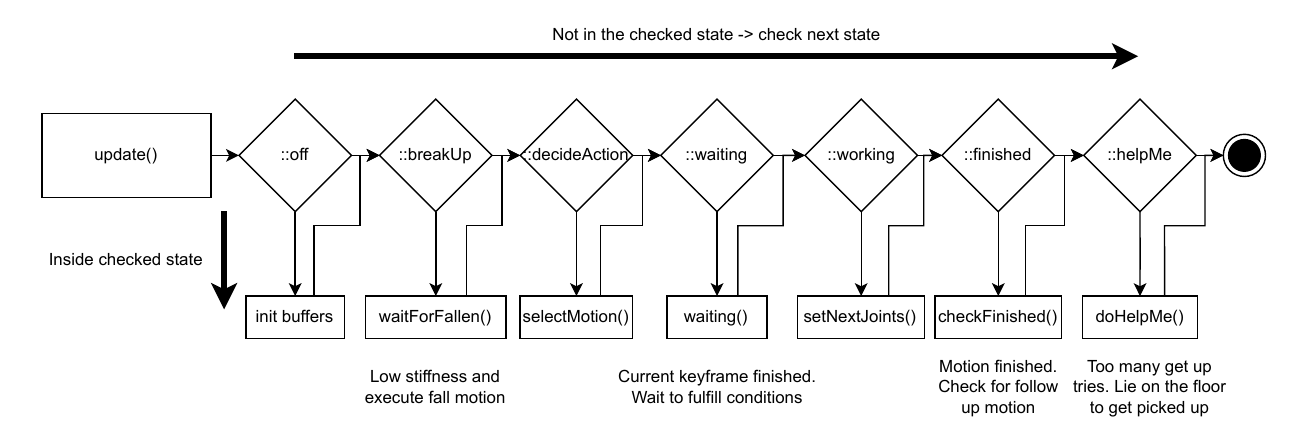}
	\caption{The overall state-based flow carried out in every execution cycle.}
	\label{f:getup:architecture}
\end{figure*}

\section{Architecture}
\label{s:architecture}

The architecture of the stand-up routine is shown in \autoref{f:getup:architecture}. The \textit{breakUp} state handles cases, in which the motion needed to stop and a fall started. Here, we ensure the robot falls safely. The \textit{decideAction} state is only used for transitions between different motions, like in the beginning to decide between the front or back stand-up, or a motion to free up a limb that is detected to be stuck. The \textit{waiting} state handles the aforementioned waiting times, in which the robot only balances but otherwise calculates no new joint positions and simply waits for the condition to fulfill. The \textit{working} state interpolates between the keyframes and applies the balancing and compensation. The \textit{finished} state signals our motion framework that the stand-up has been successfully completed. The \textit{helpMe} state is our fallback in case either the robot failed the stand-up multiple times in a row or some other defect was detected like a broken joint or a dysfunctional motor. As a dysfunctional motor is a common case that usually makes standing up impossible, this state is very important to prevent unnecessary damage caused by futile attempts to stand up.

\section{Evaluation}
\label{s:eval}

\begin{figure}[t]
    \centering
	\includegraphics[width=\columnwidth]{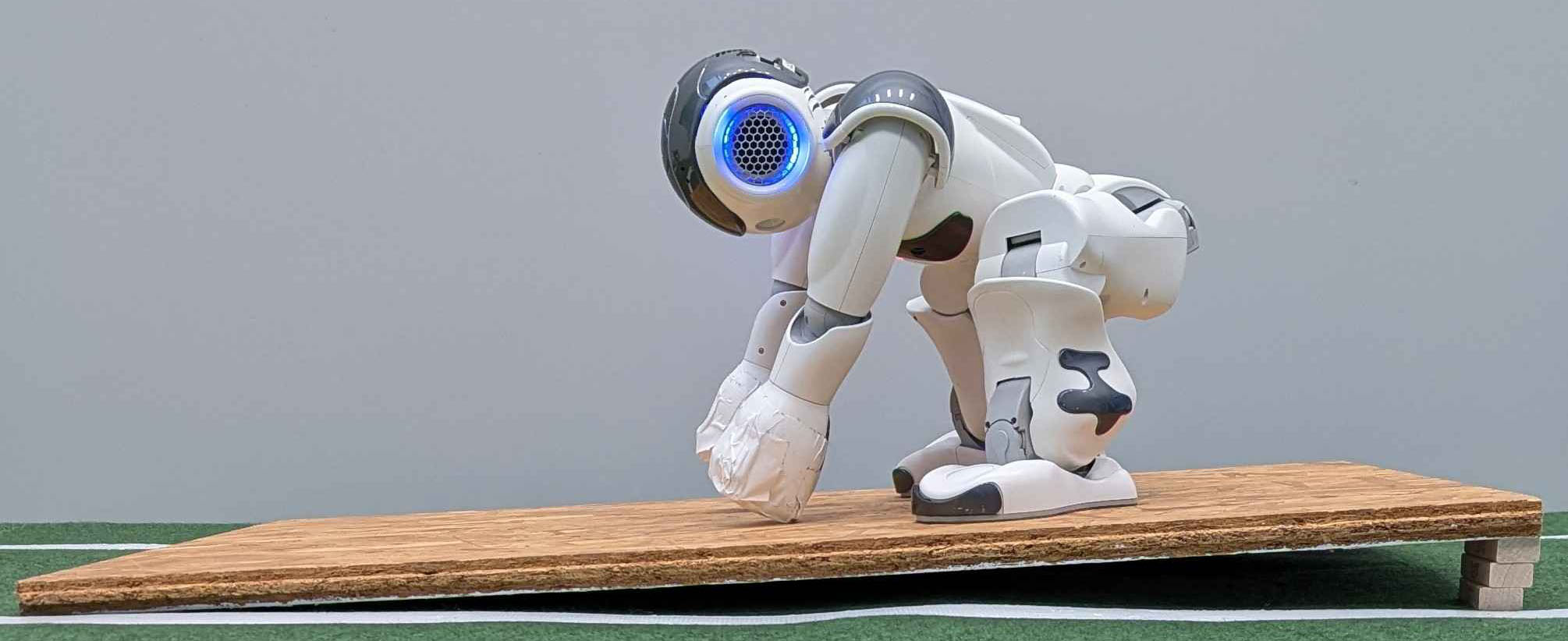}
	\caption{An example of a ramp floor with a tilt of about \qty{3}{\degree}, used to test the stand-up routines. Normally, a carpet is placed on top of the ramp for a realistic ground friction.}
	\label{f:getup:ramp}
\end{figure}

The presented stand-up approach was evaluated on different carpet conditions. Before 2019, these included a flat carpet, a flat carpet with many surface reliefs, an \qty{8}{\milli\meter} artificial turf, an artificial turf on a \qty{0.7}{\degree} ramp and a normal smooth indoor floor. Over the years, we also tested on the different carpets used at RoboCup competitions, floors on different events, as well as by increasing the tilt of the ramp condition to about \qty{3}{\degree} (see \autoref{f:getup:ramp}). We evaluated on a large set of robots with different states of wear and tear.

The approach presented in this paper has been used since 2019 on NAO V6 robots by the team B-Human and since 2021 also by other teams such as SPQR Team and RoboÉireann. The success rate and number of tries for a subset of different teams from the SPL Champions Cup is shown in \autoref{t:getUps}. The teams were selected due to their past ranking or, in the case of RoboÉireann and SPQR Team, because they are using our code release to show the success rate of our stand-up approach on previously unknown robots. The implemented code is part of the yearly B-Human code releases since 2019 and is publicly available at \cite{GitHub-BHuman-CodeRelease2024}, only missing some recent developments, which are about to become released late in 2025.

In 2019, all mentioned teams used the newly released NAO V6 robot. On such newly built robots, the influence of joint play and worn out hardware can be considered as minimal. B-Human was the only team with the presented approach and only failed two stand-up tries, while the other teams remained near the success rates of 2018 (\autoref{t:getUps:past}).

For 2022, only some parameters of the implementation were changed. However, at this point, some of the robots in use have become partially worn. Our approach, which was also used by RoboÉireann and SPQR Team, was still the best by a large margin. As RoboÉireann and SPQR Team participated remotely with robots from a shared robot pool, they mostly used the same robots while using the same code base. Therefore, it is unclear why SPQR Team performed so much worse than RoboÉireann.

\begin{table*}
	\caption{Stand-up success rate (SR) of different teams does not change significantly in the course of the years, but differs among the teams. More detailed statistics are given at \cite{FallStatistics}}\label{t:getUps}    
	\centering
	\begin{tabular}{r|c|c||c|c||c|c||c|c}
		& \multicolumn{2}{c||}{RoboCup 19} & \multicolumn{2}{c||}{GORE 22} & \multicolumn{2}{c||}{RoboCup 24} & \multicolumn{2}{c}{German Open 25}\\
		Team  & \#\,Tries & \%\,SR & \#\,Tries & \%\,SR & \#\,Tries & \%\,SR & \#\,Tries & \%\,SR \\
		\hline
		B-Human		& 81 & \textbf{97.5} & 105 & \textbf{98.1} & 180 & 81.7 & 131 & \textbf{97.7} \\
		HTWK Robots	& 153 & 73.2 & 241 & 75.5 & 741 & 64.5 & 686 & 69.1 \\
		Nao Devils 	& 211 & 73.0 & 209 & 82.8 & 502 & 76.9 & 335 & 73.4 \\
		rUNSWift	& 185 & 69.2 & 133 & 54.9 & 322 & 61.8 & & \\
		RoboÉireann\footnotemark[1]	& & & 128 & 97.7 & 354 & \textbf{91.8} & & \\
		SPQR Team\footnotemark[1]	& & & 123 & 87.0 & & & 407 & 89.9 \\
        HULKs    	& & & & & & & 536 & 83.4 \\
	\end{tabular}
\end{table*}

In 2024, the success rate decreased heavily for B-Human and also for RoboÉireann, who used our 2021 version. We cannot know the reasons for RoboÉireann, but for B-Human this large decrease is partially due to two reasons: On the one hand, most failed stand-up tries accumulated on our two most worn robots. Without them, our success rate would also be around \qty{90}{\percent}. On the other hand, most of our robots are from 2019 and very worn out, making our compensation described in \autoref{s:compensation} not being sufficient enough and sometimes repeatedly overcompensating in the same keyframe. For 2025, we changed the first part of the front stand-up, which often failed for one of our robots. Furthermore, we revised our compensation parameters to use a better subset of joints and factors, based on the position changes from the previous keyframe. This update resulted in a large improvement at the German Open 2025, while using the same robots as in 2024, including the problematic ones which only got worse in their hardware state.

\section{Conclusion and Future Work}
\label{s:conclusion}

In this paper, we presented our work on reliable stand-up motions for the humanoid robot NAO V6. We showed underlying problems, which we claim to effect other teams as well. We proposed a novel idea on how to handle errors in the execution of motor position commands to still successfully execute the requested stand-up motions. Our idea of applying the errors directly on other joints is quite straightforward yet effective in handling  stabilization problems. This is of particular relevance for the many old robots, which are in different states of wear and tear.
We observed that all teams in the Standard Platform League struggle with such motions. As of today, our approach shows the best results, as shown by an analysis of several teams during multiple official competitions. 

Our approach can still be refined, as the compensation is currently parameterized by hand. The human work was done by some algorithmic approach, which could be automated, as most of its observed problems were due to overcompensation. In addition, the keyframes used as a baseline could be further improved or replaced entirely to reduce the occurrence of problems like stuck joints, which are a consequence of the keyframe approach. This could be achieved, for instance, by generating reference values based on the current observed state of the robot.

Furthermore, due to the latest research advancements and the newest open source projects, the usage of IsaacGym to learn policies for standing up appears to be a promising approach and will be investigated in future work.

\footnotetext[1]{Teams that use our proposed stand-up routine.}
\printbibliography

\end{document}